\documentclass{ieeeaccess}

\usepackage{amsmath,amsfonts}
\usepackage{array}
\usepackage{textcomp}
\usepackage{stfloats}
\usepackage{url}
\usepackage{verbatim}
\usepackage{cite}

\usepackage{amssymb}

\usepackage[ruled,vlined]{algorithm2e}
\usepackage{hyperref}
\usepackage{float}
\usepackage{multirow}
\usepackage{booktabs,caption}
\usepackage[flushleft]{threeparttable}
\usepackage{subcaption}
\floatstyle{plaintop}
\restylefloat{table}

\usepackage{stackengine}
\usepackage{arydshln}
\usepackage[pdftex]{graphicx}

\usepackage{fancyhdr}
\usepackage{kantlipsum}
\usepackage{eso-pic}

\def\BibTeX{{\rm B\kern-.05em{\sc i\kern-.025em b}\kern-.08em
    T\kern-.1667em\lower.7ex\hbox{E}\kern-.125emX}}

\usepackage{soul} 
\usepackage{color} 
\DeclareRobustCommand{\rev}[1]{\hl{#1}} 
\DeclareRobustCommand{\rev}[1]{#1} 

\soulregister\cite7
\soulregister\ref7
\soulregister\textbf7

\usepackage[font=it]{caption} 

\begin{document}

\AddToShipoutPictureBG*{%
	\AtPageLowerLeft{%
		\setlength\unitlength{1in}%
		\hspace*{\dimexpr0.5\paperwidth\relax}
		\makebox(0,0.63)[c]{2169-3536~\copyright2025 IEEE. This work has been accepted for publication in IEEE Access.}
		\makebox(0,0.3)[c]{The published version can be accessed at \href{https://ieeexplore.ieee.org/document/11313052}{https://ieeexplore.ieee.org/document/11313052}. DOI: \href{https://doi.org/10.1109/ACCESS.2025.3647530}{10.1109/ACCESS.2025.3647530}}
}}

\history{Date of publication xxxx 00, 0000, date of current version xxxx 00, 0000.}
\doi{10.1109/ACCESS.2024.0429000}

\title{DeepIPCv2: LiDAR-powered Robust Environmental Perception and Navigational Control for Autonomous Vehicle}
\author{\uppercase{Oskar Natan}\authorrefmark{1}, \IEEEmembership{Member, IEEE} and
\uppercase{Jun Miura}\authorrefmark{2},
\IEEEmembership{Member, IEEE}}

\address[1]{Department of Computer Science and Electronics, Universitas Gadjah Mada, Yogyakarta 55281 Indonesia (e-mail: oskarnatan@ugm.ac.id)}
\address[2]{Department of Computer Science and Engineering, Toyohashi University of Technology, Aichi 441-8580 Japan (e-mail: jun.miura@tut.jp)}


\markboth
{{\bf O. Natan and J. Miura}: DeepIPCv2: LiDAR-powered Robust Environmental Perception and Navigational Control for Autonomous Vehicle}
{{\bf O. Natan and J. Miura}: DeepIPCv2: LiDAR-powered Robust Environmental Perception and Navigational Control for Autonomous Vehicle}

\corresp{Corresponding author: Oskar Natan (e-mail: oskarnatan@ugm.ac.id).}

\begin{abstract}
We propose DeepIPCv2, an end-to-end autonomous driving framework that integrates LiDAR-based environmental perception with command-specific control learning. Unlike prior camera-reliant models, DeepIPCv2 employs point cloud segmentation and multi-view projection (front and bird’s-eye views) to construct robust scene representations. These features are fused and decoded through a combination of gated recurrent units (GRU), command-specific multi-layer perceptrons (MLP), and PID controllers to estimate both waypoints and navigational control commands. This design enhances maneuverability and addresses action imbalance in driving datasets. To validate the model, we constructed a dataset covering diverse illumination conditions and conducted ablation studies and comparative tests against recent methods, including TransFuser. Results demonstrate that DeepIPCv2 achieves the lowest total metric error and the fewest driving interventions, highlighting both its robustness to illumination changes and its improved control accuracy. By releasing the codes at \href{https://github.com/oskarnatan/DeepIPCv2}{https://github.com/oskarnatan/DeepIPCv2} later, we aim to support reproducibility and future advancements in end-to-end autonomous driving research.
\end{abstract}


\titlepgskip=-21pt

\maketitle

\section{Introduction}\label{sec:intro}

\begin{figure}[t]
	\begin{center}
		\includegraphics[width=\linewidth]{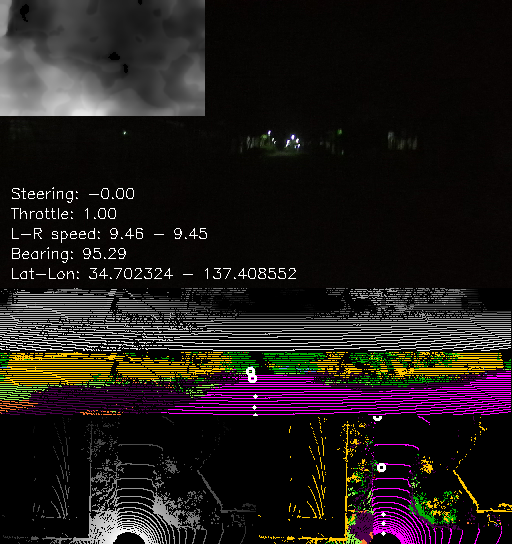}
		
	\end{center}
	\caption{DeepIPCv2 perceives the environment by encoding a set of segmented point clouds that are projected into front and top-view perspectives. Then, the extracted features are used to estimate waypoints (white dots) and navigational controls to drive the vehicle following the route points (white hollow circles). Meanwhile, the RGBD image is only for record purposes. It shows how the RGBD camera fails in capturing surrounding information as it cannot provide a clearly visible set of RGB image and depth map. Therefore, DeepIPCv2 employs a LiDAR sensor which has its own lasers as the light source to provide point clouds that are not affected by poor illumination conditions.}
	\label{fig:overview}
\end{figure}

Perceiving environment has always been a crucial stage in navigation systems as it is important to understand the surrounding before making any decisions \cite{av_vision0}\cite{avisionx}\cite{avisiony}\cite{avisionz}. A model can perceive the environment by performing many kinds of vision tasks such as semantic segmentation and depth estimation \cite{obj_det}\cite{sem_seg}\cite{matsuzaki}\cite{segx}\cite{dep_est}. To achieve a better performance, a lot of works have been proposed to improve scene understanding capability such as feeding a sequence of RGB images \cite{seg_seq}, training with multi-task learning paradigm \cite{mtl}\cite{mtl_tii}, and using sensor fusion techniques \cite{sensor_fuse0} such as combining RGB images with dynamic vision sensor (DVS) images \cite{rgbdvs}, depth (RGBD) images \cite{rgb_d}, or even radar \cite{senfus_tie}. However, since these approaches rely on cameras, their performance may decrease or even fail under poor illumination conditions where everything is not clearly visible. One example of a camera-powered model for autonomous driving is our previous work namely DeepIPC (Deeply Integrated Perception and Control) \cite{deepipc}. Concisely, DeepIPC perceives the environment based on the captured RGBD images and drives the vehicle following a set of route points. However, its drivability decrease under low-light conditions as the camera is heavily affected by the illumination changes. Thus, an alternative sensor that is robust against illumination changes such as LiDAR is needed to perform a stable observation \cite{lidar1}\cite{lidar3}\cite{lidar4}.

In this research, we introduce DeepIPCv2, an improved version of DeepIPC \cite{deepipc} and our simulation work \cite{oskar_tiv}, where the focus of improvement is to tackle the challenge of driving under poor illumination conditions. As shown in Fig. \ref{fig:overview}, the RGBD camera cannot provide a clear visible set of RGB image and depth map. Hence, DeepIPCv2 uses a LiDAR sensor and employs a point cloud segmentation model to perceive the environment. This enables better reasoning as the model can distinguish traversable and non-traversable areas easily and avoid collision by knowing the existence of other objects around the ego vehicle. By encoding these point clouds, the perception module can provide stable and better features to the controller module for estimating waypoints and navigational control. Thus, DeepIPCv2 can maintain its drivability performance even when driving at night. In addition, we also modify the controller module by adding a set of command-specific multi-layer perceptrons (MLP) to assist the PID controllers, improving the maneuverability. The novelties of this work can be summarized as follows:

\begin{itemize}
	\item \rev{\textbf{LiDAR-based Semantic Multi-View Perception:} Unlike density-based projections (e.g., TransFuser \cite{transfuser2}), we employ a semantic projection pipeline. We utilize PolarNet \cite{polarnet} for efficient point cloud segmentation and project the results into Front-View and Bird’s-Eye-View (BEV). This provides the controller with explicit class-aware traversability information, ensuring robustness in low-light conditions where cameras fail.}
	
	\item \rev{\textbf{Hybrid Control with Command-Specific Decoders:} We introduce a controller integrating a Gated Recurrent Unit (GRU) with command-specific Multi-Layer Perceptrons (MLPs) and PID controllers. This hybrid design addresses the action imbalance in driving datasets and improves maneuverability by dedicating decoders to specific high-level commands.}
	
	\item \textbf{Open Research and Future Advancements:} To contribute to ongoing research and future advancements, we will release our codes and datasets, promoting transparency and collaboration in autonomous driving technology.
	Link: \href{https://github.com/oskarnatan/DeepIPCv2}{https://github.com/oskarnatan/DeepIPCv2} and \href{https://youtu.be/IsZ1HP5QjWc}{https://youtu.be/IsZ1HP5QjWc}
\end{itemize}

\section{Related Work}
In this section, we review some works that discuss point cloud processing. Then, we provide reviews on some notable works that use a LiDAR sensor for autonomous driving. 

\subsection{LiDAR-powered Perception} \label{subs:perception}

LiDAR is a sensor that is considered to be more robust than an RGBD camera when dealing with poor illumination conditions. Unlike RGBD images, the point clouds are not affected by the illumination changes since the LiDAR has its own lasers as the light source to observe the environment \cite{robust_lidar}\cite{lidar2}\cite{lidar_tie}. Furthermore, together with plenty of point cloud segmentation models and projection techniques, many kinds of data representations can be formed to provide meaningful information \cite{pt_cloud_av0}\cite{pt_cloud_av1}\cite{lidar_tii}. In this research, we employ a point cloud segmentation model to achieve a robust environmental perception and scene understanding.

To date, there are plenty of works in the development of point cloud segmentation models. In the Semantic KITTI dataset \cite{kitti}, the current state-of-the-art is achieved by a model named 2DPASS \cite{2dpass}. However, its performance needs to be justified further in a very poor illumination condition as this model uses RGB images to assist the segmentation process. A point cloud segmentation model that only uses a LiDAR is proposed by Hou et. al. \cite{pvkd} which is currently the runner-up in the semantic point cloud segmentation challenge. Although it has a great performance, its size and latency are not suitable for a device with limited computation power. For deployment purposes, we need to consider the trade-off between speed and performance. Since we also seek robustness, the model must only use LiDAR in performing point cloud segmentation. Thus, we select PolarNet \cite{polarnet}, a lightweight point cloud segmentation model that offers an optimal trade-off between accuracy and latency for embedded applications. \rev{We also acknowledge recent advancements such as PC-BEV \cite{pcbev}, which offers efficient fusion frameworks, though we prioritize the established stability of PolarNet.}

\subsection{End-to-end Model} \label{subs:e2e}

%
With the rapid deep learning research, perception and control parts can be coupled together in an end-to-end manner to avoid manual integration that can lead to information loss. An end-to-end model is proven to have a better generalization as it can leverage the feature-sharing mechanism within its layers \cite{e2e_av1}\cite{e2e_ava}. Moreover, each neuron can receive extra supervision from a multi-task loss formula that considers multiple performance criteria \cite{ishihara}\cite{mtl_a}. This results in a compact model that is relatively small but has a great performance which is preferable for real deployment.

Recent progress is made by Chitta et. al. \cite{neat} where a camera-powered end-to-end model is deployed to perform automated driving in a simulated environment. The RGB encoder of this model is guided by bird's eye-view (BEV) semantic prediction to provide better features to the controller decoder. Although its performance in poor illumination conditions is promising, this model is practically hard to train as it is difficult to create BEV semantic ground truth in a real dataset. Then, a different work is proposed by Prakash et. al. where a camera-LiDAR fusion model named TransFuser \cite{transfuser}\cite{transfuser2} is deployed to handle various scenarios in autonomous driving. The camera is used to capture an RGB image in front of the vehicle, while the LiDAR is used to capture point clouds around the vehicle. The point clouds are projected into a 2-bin histogram over a 2D BEV grid with a fixed resolution \cite{2bin_transfuser}. With this configuration, the model can perceive from both front and BEV perspectives. Then, a certain transformer-based module is used to learn the relation between the RGB image and the projected point clouds to achieve a better perception. Considering its performance and robustness, we use TransFuser and its variants for comparative study purposes.

\section{Methodology}

In this section, we describe the architecture of DeepIPCv2 and some metrics used to train and evaluate the model. Then, we also explain how the data is collected to train, validate, and test the model.

\begin{figure*}
	\begin{center}
		\includegraphics[width=\linewidth]{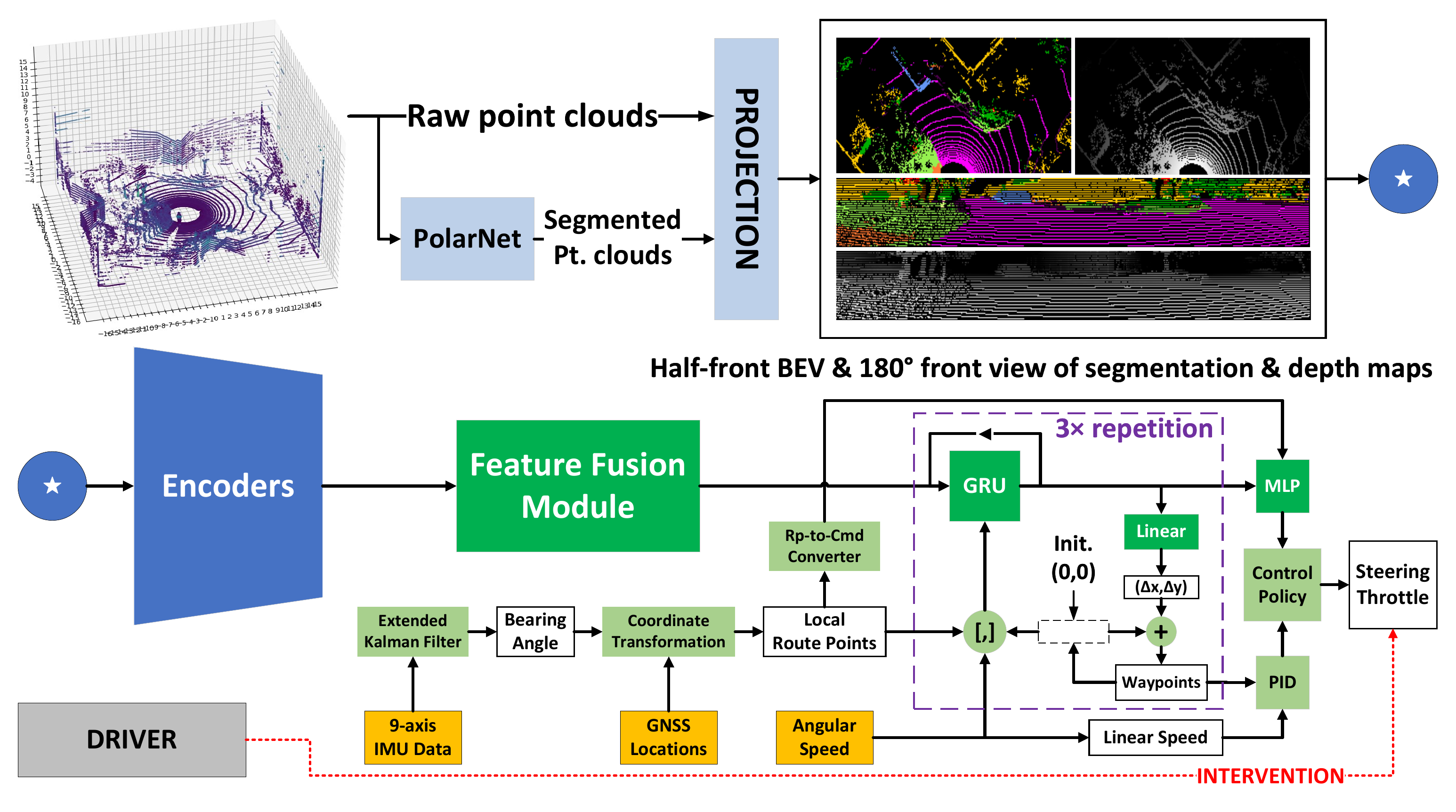}
		
	\end{center}
	\vspace{-3mm}
	\caption{The architecture of DeepIPCv2. The blue and green blocks are the perception and controller modules respectively. Darker blocks are trainable, while light-colored blocks are not. In the perception module, PolarNet \cite{polarnet} is employed to support point cloud segmentation. Then, the architecture of encoders and feature fusion modules can be seen in Fig. \ref{fig:model2}.}
	\label{fig:model}
\end{figure*}

\begin{figure*}
	\begin{center}
		\includegraphics[width=\linewidth]{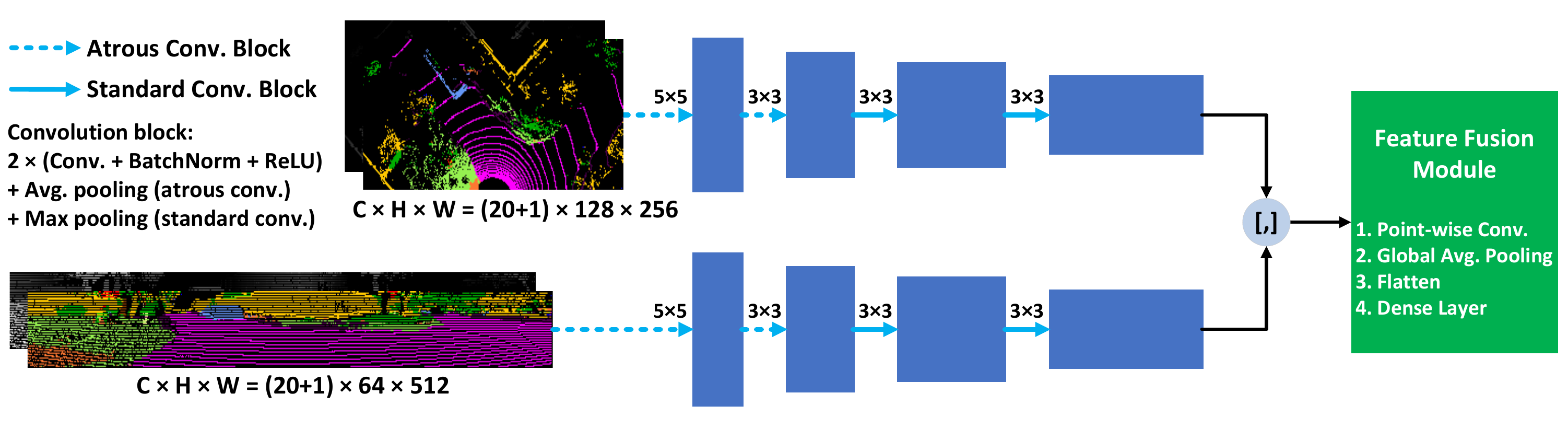}
		
	\end{center}
	\caption{The architecture of the encoders and the feature fusion module. We use atrous convolution blocks \cite{atrous_conv} with different kernel sizes and dilation rates to capture low-level features from the projected point clouds that have vacant regions. Then, both top and front features are fused together and their relationship are learned by the feature fusion module.}
	\label{fig:model2}
\end{figure*}

\subsection{Proposed Model} \label{subs:model}

\subsubsection{Perception Phase}

Similar to DeepIPC \cite{deepipc}, DeepIPCv2 is also a model that handles perception and control tasks simultaneously. However, unlike DeepIPC which takes an RGBD image, DeepIPCv2 takes a set of LiDAR point clouds to perceive the environment. Since LiDAR is not affected by poor illumination conditions, the perception module becomes more robust and can provide stable features to the controller module. Thus, the model can estimate waypoints and navigation control properly even when driving at night. As shown in Fig. \ref{fig:model}, DeepIPCv2 employs PolarNet \cite{polarnet}, a light-weight point cloud segmentation model pre-trained on the Semantic KITTI dataset \cite{kitti} to segment LiDAR point clouds into twenty object classes as mentioned in Table \ref{tab:data_info}. \rev{By leveraging these pre-trained weights, our system generates robust semantic "pseudo-labels" that serve as stable high-level features for the controller. This design ensures that the perception quality is not constrained by the specific object class distribution of our smaller campus dataset.} Based on our previous work \cite{oskar_tiv}, perceiving the environment from more perspectives can improve perception and lead to better drivability. Hence, we project segmented point clouds to form one hot-encoded image-like array that shows front-view and bird's eye-view (BEV) perspectives of the surrounding area. Each array is expressed as $\mathbb{R} \in \{0,1\}^{C \times H \times W}$, where $H \times W$ is the spatial dimension with $H \times W = 64 \times 512$ for the front-view array and $H \times W = 128 \times 256$ for the BEV array. Meanwhile, $C = 21$ represents the number of channels that are responsible for twenty object classes and a logarithmic depth of the point clouds. In forming the BEV array, we consider an area of 16 meters to the front, left, and right. Meanwhile, for the front-view array, we consider all point clouds in front of the vehicle forming a 180-degree field of view. The projection follows a spatial quantization scheme defined by the sensor's field of view and the desired resolution.

\paragraph{Bird’s-Eye View Projection}

Let each point be denoted as $\mathbf{p}_i = (x_i, y_i, z_i, c_i)$, where $(x_i, y_i, z_i) \in \mathbb{R}^3$ is the point coordinate in the ego-centric frame and $c_i \in \{1, ..., C{-}1\}$ is the semantic class index. Given a BEV grid of size $H_b \times W_b$ covering the range $x \in [-x_{\text{max}}, x_{\text{max}}]$ and $y \in [0, y_{\text{max}}]$, the 2D grid indices $(u, v)$ are computed with \ref{eq:proj_1}.

\begin{equation}\label{eq:proj_1}
	u = \left\lfloor \frac{x_i + x_{\text{max}}}{2x_{\text{max}}} \cdot H_b \right\rfloor, \quad
	v = \left\lfloor \frac{y_i}{y_{\text{max}}} \cdot W_b \right\rfloor
\end{equation}

The corresponding values in the BEV tensor are assigned with (\ref{eq:proj_2}) and (\ref{eq:proj_3}) where \rev{$\epsilon = 10^{-6}$ is a small constant added to prevent logarithmic singularity ($\log(0)$) when points are extremely close to the sensor center. We utilize a logarithmic scale for depth to compress the high dynamic range of LiDAR (0--100m), preserving feature gradients for nearby objects.}

\begin{equation}\label{eq:proj_2}
	\mathbf{R}_{c_i, u, v} = 1
\end{equation}

\begin{equation}\label{eq:proj_3}
	\mathbf{R}_{C, u, v} = \log\left( \sqrt{x_i^2 + y_i^2 + z_i^2} + \epsilon \right)
\end{equation}

\paragraph{Front-View Projection}

Each point is also mapped into angular coordinates for front-view projection with formula \ref{eq:proj_4}.

\begin{equation}\label{eq:proj_4}
	\theta_i = \tan^{-1} \left( \frac{y_i}{x_i} \right), \quad
	\phi_i = \tan^{-1} \left( \frac{z_i}{\sqrt{x_i^2 + y_i^2}} \right)
\end{equation}

Assuming a horizontal field of view $[\theta_{\min}, \theta_{\max}]$ and vertical field of view $[\phi_{\min}, \phi_{\max}]$ for a front-view grid of size $H_f \times W_f$, the pixel locations are given by \ref{eq:proj_5} and the front-view tensor is populated similarly with \ref{eq:proj_6}.

\begin{equation} \label{eq:proj_5}
	u = \left\lfloor \frac{\phi_i - \phi_{\min}}{\phi_{\max} - \phi_{\min}} \cdot H_f \right\rfloor, \quad
	v = \left\lfloor \frac{\theta_i - \theta_{\min}}{\theta_{\max} - \theta_{\min}} \cdot W_f \right\rfloor
\end{equation}

\begin{equation} \label{eq:proj_6}
	\mathbf{R}_{c_i, u, v} = 1, \quad
	\mathbf{R}_{C, u, v} = \log\left( \sqrt{x_i^2 + y_i^2 + z_i^2} + \epsilon \right)
\end{equation}

This dual-perspective projection allows DeepIPCv2 to capture both semantic context and geometric structure, improving robustness in navigational control estimation. To process these projections, we use two different encoders that are made of atrous and standard convolution blocks as shown in Fig. \ref{fig:model2}. Atrous convolution blocks \cite{atrous_conv} are used to deal with some vacant regions in the projected LiDAR point clouds at the early encoding process. As the kernel sizes and dilation rates can be adjusted, an atrous convolution layer is more suitable than a standard convolution layer for extracting the features. Then, we also configure the pooling size after each convolution block to match the output size of both encoders. With this configuration, DeepIPCv2 has a better scene understanding capability as it can perceive from two different perspectives that clearly show traversable and non-traversable regions.

\rev{Later, we perform a series of ablation studies to systematically assess the contribution of each input modality and perception configuration. First, we construct two simplified model variants: one that uses only the logarithmic-depth point clouds and another that relies solely on the segmented point clouds. Evaluating these variants allows us to understand the relative importance of each representation. Then, we further extend the ablation process by comparing this variant against models that perceive the environment from a single viewpoint, either exclusively from the front perspective or from the bird’s-eye view (BEV). This additional comparison is crucial for demonstrating the necessity of multi-view perception.}

\subsubsection{Control Phase}

The control phase begins by fusing \rev{both high-level perception features} to produce a latent space composed of 192 feature elements that encapsulate the information of the surrounding based on two perspectives of view. This process is done by the feature fusion module that consists of a point-wise convolution layer, a global average pooling layer, and a dense layer. \rev{"Feature Fusion" here refers to the concatenation of the flattened feature maps from both the Front-View and BEV encoders.} Then, we use the first and second route points, the left and right wheel's angular speed, and predicted waypoints to bias the latent space in the gated recurrent unit (GRU) \cite{gru}.

\rev{Finally, the biased latent space is decoded by: 1) A set of Command-Specific MLPs to estimate navigational control (Steering $\delta$, Throttle $\tau$) directly. 2) Two linear layers to predict waypoints that will be translated into navigational control by a set of two PID controllers. A "waypoint" is defined as a target coordinate pair $(x,y)$ in the vehicle's local coordinate frame representing the desired future path. The PID controllers process the waypoints to generate a parallel control signal. We utilized a standard PID formulation where the parameters were tuned experimentally using the Ziegler-Nichols method to ensure stability without oscillation. The values used are $K_p=1.00$, $K_i=0.25$, and $K_d=0.15$ for steering, and $K_p=2.50$, $K_i=0.25$, $K_d=0.50$ for throttle.} To be noted, both MLP and PID controllers assume the model of the robotic vehicle as a nonholonomic unicycle since it has motorized rear wheels and omnidirectional front wheels. Thus, it cannot perform translational movement on the lateral axis, but it can move only along its longitudinal axis (forward and backward) and can rotate around a vertical axis passing through its center. As shown in Fig. \ref{fig:model}, the process inside the purple box is looped three times as DeepIPCv2 predicts three waypoints. Two linear layers are used to predict $\Delta x$ and $\Delta y$, a gap between the current waypoint and the next waypoint. Thus, the exact coordinate of the next waypoint can be calculated with (\ref{eq:wp}).

\begin{equation} \label{eq:wp}
	x_{i+1}, y_{i+1} = (x_{i}+\Delta x), (y_{i}+\Delta y)
\end{equation}

\begin{algorithm}[t] 
	\SetAlgoLined
	$\Theta = \frac{Wp_1+Wp_2}{2}$; $\theta = \tan^{-1}\big(\frac{\Theta[1]}{\Theta[0]}\big)$\\
	$\gamma = 1.75\times||Wp_1 - Wp_2||_F$; $\nu = \frac{(\omega_l + \omega_r)}{2} \times r$\\
	\dotfill\\
	\uIf{$Rp^x_1 \leq -4m$ or $Rp^x_2 \leq -8m$}{
		$Cmd = 2$ (turn right)
	}
	\uElseIf{$Rp^x_1 \geq 4m$ or $Rp^x_2 \geq 8m$}{
		$Cmd = 1$ (turn left)
	}
	\uElse{
		$Cmd = 0$ (go straight)
	}
	\dotfill\\
	$\mathbf{MLP}_{\{ST,TH\}} = \mathbf{MLP}^{Cmd}(\mathcal{Z})$\\
	$\mathbf{PID}_{\{ST,TH\}} = \mathbf{PID}^{Lat}(\theta-90), \mathbf{PID}^{Lon}(\gamma-\nu)$\\
	\uIf{$\mathbf{MLP}_{TH} \geq 0.1$ and $\mathbf{PID}_{TH} \geq 0.1$}{
		\uIf{$|\mathbf{MLP}_{ST}| \geq 0.1$ and $|\mathbf{PID}_{ST}| < 0.1$}{
			steering $= \mathbf{MLP}_{ST}$\\
		}
		\uIf{$|\mathbf{MLP}_{ST}| < 0.1$ and $|\mathbf{PID}_{ST}| \geq 0.1$}{
			steering $= \mathbf{PID}_{ST}$\\
		}
		\uElse{
			steering $= \beta_{00} \mathbf{MLP}_{ST} + \beta_{10} \mathbf{PID}_{ST}$\\
		}
		throttle $= \beta_{01} \mathbf{MLP}_{TH} + \beta_{11} \mathbf{PID}_{TH}$\\
	}
	\uElseIf{$\mathbf{MLP}_{TH} \geq 0.1$ and $\mathbf{PID}_{TH} < 0.1$}{
		steering $= \mathbf{MLP}_{ST}$; throttle $= \mathbf{MLP}_{TH}$\\
	}
	\uElseIf{$\mathbf{MLP}_{TH} < 0.1$ and $\mathbf{PID}_{TH} \geq 0.1$}{
		steering $= \mathbf{PID}_{ST}$; throttle $= \mathbf{PID}_{TH}$\\
	}
	\uElse{
		steering $= 0$; throttle $= 0$\\
	}
	\dotfill\\
	\caption{Control Policy}
	\label{alg:control}
	\begin{tablenotes}
		\small
		
		\item \rev{ $Rp^x_{\{1,2\}}$: route point's $x$ position in the local coordinate}
		\item \rev{ $Wp_{\{1,2\}}$: first and second waypoints}
		\item \rev{ $\mathcal{Z}$: GRU's latent space}
		\item \rev{ $\omega_{\{l,r\}}$: left/right angular speed (rad/s)}
		\item \rev{ $r$: vehicle's rear wheel radius (0.15 m)}
		\item \rev{ $\Theta$: aim point, a middle point between $Wp_1$ and $Wp_2$}
		\item \rev{ $\theta$: heading angle derived from the aim point $\Theta$}
		\item \rev{ $\gamma$: desired speed, 1.75 $\times$ Frobenius norm of $Wp_1$ and $Wp_2$}
		\item \rev{ $\nu$: linear speed (m/s), the mean of $\omega_l$ and $\omega_r$ multiplied by $r$ 	}
		\item \rev{ $\beta\in\{0,...,1\}^{2\times2}$ is a set of control weights} initialized with:
		\item $\beta_{00} = \frac{\alpha_1}{\alpha_1+\alpha_0}$; $\beta_{10} = 1 - \beta_{00}$; $\beta_{01} = \frac{\alpha_2}{\alpha_2+\alpha_0}$; $\beta_{11} = 1 - \beta_{01}$
		\item where $\alpha_0, \alpha_1, \alpha_2$ are loss weights computed by MGN
		\item algorithm \cite{mgn} (see Subsection \ref{subs:train} for more details)

	\end{tablenotes}
\end{algorithm}

To predict the first waypoint, the current waypoint is initialized with the vehicle position in the local coordinate which is always at (0,0). Then, the waypoints are processed by two PID controllers to produce a set of navigational control consisting of steering and throttle levels. Besides using PID controllers, DeepIPCv2 also predicts navigational control directly by decoding biased latent using MLP. However, unlike DeepIPC which employs only one MLP, DeepIPCv2 employs a set of command-specific MLPs for better maneuverability as demonstrated by Huang et. al \cite{huang_model}. Each of the command-specific MLPs act as a task-specific decoder that receives the same features from the same encoder. Then, since each decoder treats each action (turn left, turn right, or go straight) independently, the model has better maneuverability as it has more focus by deploying a dedicated MLP for each action. Moreover, this configuration can also deal with the imbalance number of actions in the driving records (e.g. the number of observation sets for go straight is larger than turn left or turn right). Meanwhile, the commands are generated automatically based on the route point's $x$ position. The rule that generates the command and the policy which outputs the final action is summarized on Algorithm \ref{alg:control}. \rev{The command thresholds ($\pm$4m and $\pm$8m) were determined empirically based on the average road width of the campus environment to ensure the vehicle remains centered during navigation tasks.}

\rev{To enable end-to-end learning, the model is optimized using a Multi-Task Loss ($\mathcal{L}_{MTL}$) defined as:}
\begin{equation} \label{eq:mtlloss_moved}
	\mathcal{L}_{MTL} = \alpha_0\mathcal{L}_{WP} + \alpha_1\mathcal{L}_{ST} + \alpha_2\mathcal{L}_{TH},
\end{equation}
\rev{where $\mathcal{L}_{WP}$ is the L1 loss for waypoints, and $\mathcal{L}_{ST/TH}$ are L1 losses for steering and throttle. $\alpha$ terms are dynamic weights computed via Modified Gradient Normalization (MGN) \cite{mgn}.}

\subsubsection{Global-to-Local Coordinate Transformation}

Furthermore, other measurement quantities and formulas are needed to transform the route points from the global GNSS coordinate to the local coordinate where the vehicle is always positioned at (0,0). To obtain the local coordinate for each route point $i$, the relative distance $\Delta x_{i}$ and $\Delta y_{i}$ between vehicle location $Ro$ and route point location $Rp_i$ need to be calculated first. Using the information of global longitude-latitude given by the GNSS receiver, the relative distance can be calculated with (\ref{eq:lon_meter}) and (\ref{eq:lat_meter}).

\begin{equation} \label{eq:lon_meter}
	\Delta x_{i} = (Rp^{Lon}_i - Ro^{Lon}) \times \frac{\mathcal{C}_e \times \cos(Ro^{Lat})}{360},
\end{equation}
\begin{equation} \label{eq:lat_meter}
	\Delta y_{i} = (Rp^{Lat}_i - Ro^{Lat}) \times \frac{\mathcal{C}_m}{360},
\end{equation}

\noindent where $\mathcal{C}_e$ and $\mathcal{C}_m$ are earth's equatorial and meridional circumferences which are 40,075 and 40,008 km, respectively. Then, the local coordinate of each route point $Rp^{(x,y)}_i$ can be obtained by applying a rotation matrix as in (\ref{eq:bev_transform}).

\begin{equation} \label{eq:bev_transform}
	\begin{bmatrix}
		Rp^{x}_i \\
		Rp^{y}_i 
	\end{bmatrix} = 
	\begin{bmatrix}
		\cos(\theta_{ro}) & -\sin(\theta_{ro}) \\
		\sin(\theta_{ro}) & \cos(\theta_{ro})
	\end{bmatrix}^T
	\begin{bmatrix}
		\Delta x_{i}\\
		\Delta y_{i}
	\end{bmatrix},
\end{equation}

\noindent where $\theta_{ro}$ is the vehicle's absolute orientation to the north pole (bearing angle). In this research, the bearing angle is estimated by the \rev{Extended Kalman Filter} (EKF) based on the measurement of 3-axial acceleration, angular speed, and magnetic field retrieved from a 9-axis IMU sensor. To be noted, due to the GNSS inaccuracy and noisy IMU measurements the global-to-local transformation may not be perfect. Thus, the model is expected to learn implicitly how to compensate for this issue during the training process.

\subsection{Dataset}\label{subs:dataset}

\begin{figure}[t]
	\begin{center}
		\includegraphics[width=\linewidth]{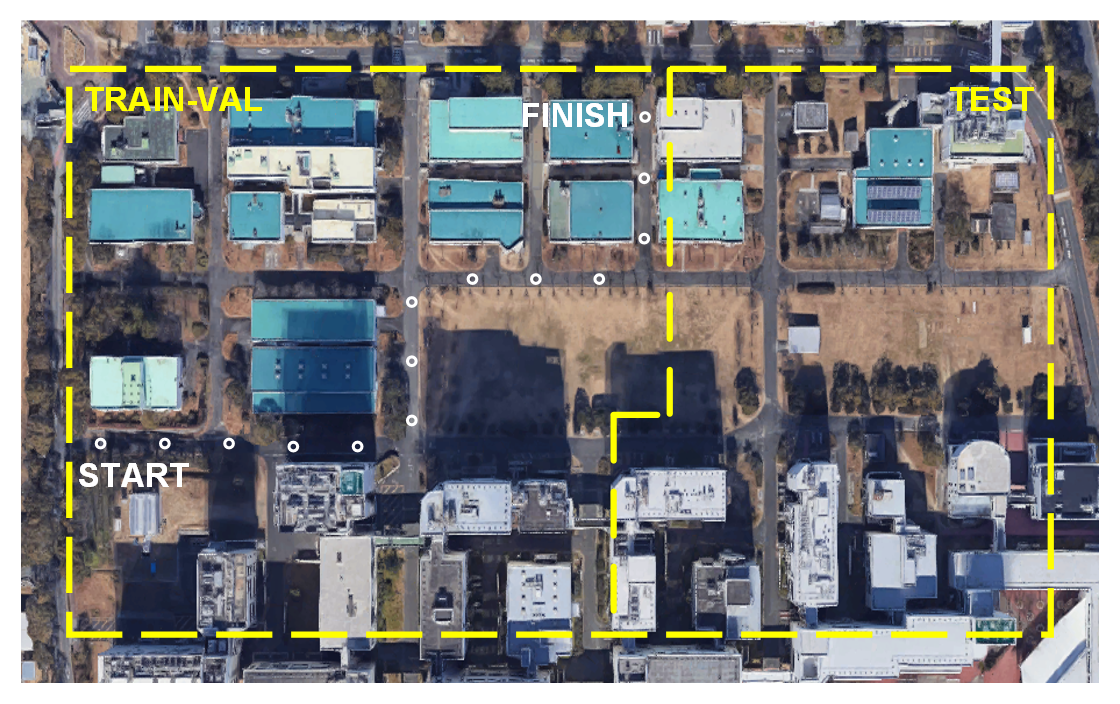}
		
	\end{center}
	\caption{The experiment areas (more detail at \href{https://goo.gl/maps/9rXobdhP3VYdjXn48}{https://goo.gl/maps/9rXobdhP3VYdjXn48}). White hollow circles represent route points.} 
	\label{fig:tut_map}
\end{figure}

\begin{figure}[t]
	\begin{center}
		\includegraphics[width=0.9\linewidth]{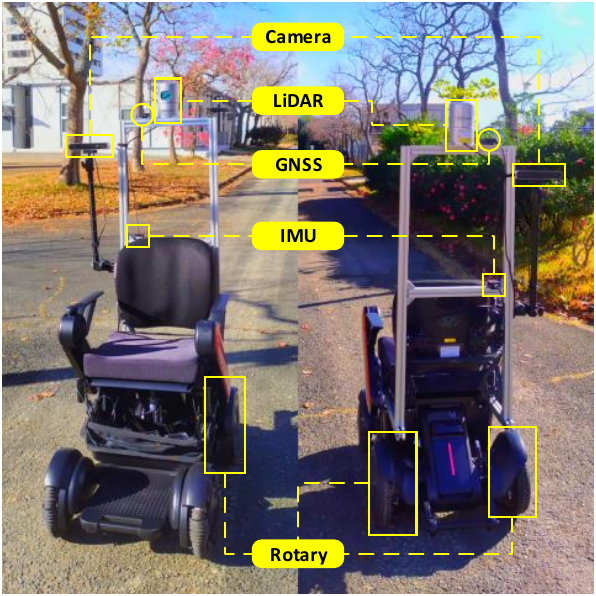}
		
	\end{center}
	\caption{Sensor placement on the robotic vehicle. The vehicle is categorized as a nonholonomic unicycle since it has motorized rear wheels and omnidirectional front wheels.} 
	\label{fig:vehicle}
\end{figure}

A dataset that consists of expert driving records is needed for behavior cloning \cite{imit4}\cite{imita}\cite{hidehito}\cite{imitb}. To create the dataset for training, validation, and test (train-val-test), we record observation data while driving the robotic vehicle at a speed of 1.25 m/s in an area inside Toyohashi University of Technology, Japan as shown in Fig. \ref{fig:tut_map}. \rev{This speed limit is imposed by the hardware constraints of the WHILL Model C2 platform and ensures safety during data collection in the pedestrian-populated campus environment.} We record the driving data at noon, in the evening, and at night to vary the experiment conditions. In the train-val area, there are 12 different routes where the driving data is recorded one time for each condition. Meanwhile, in the test area, there are 6 different routes where the driving data is recorded three times for each condition. Each route has a set of route points with 12 meters gap that shows the path to the finish point. The vehicle must follow this path to complete the route.

Recorded at a rate of 4 Hz, one sample of observation data is composed of a set of LiDAR point clouds, GNSS latitude-longitude, 9-axis IMU measurement, left and right wheel's angular speeds, and the level of steering and throttle. We also record the RGB image which is used by another model for comparison. Then, as the ground truth for the waypoints prediction task, we use the vehicle's trajectory location in one second, two seconds, and three seconds in the future (relative to the vehicle's current location at the current time). The trajectory is estimated by a built-in IMU-based odometry algorithm embedded in the robotic vehicle. Meanwhile, as the ground truth for the navigational control estimation task, we use the record of steering and control levels at the time. The devices used to retrieve observation data are mentioned in Table \ref{tab:data_info}. Meanwhile, how they are mounted on the vehicle can be seen in Fig. \ref{fig:vehicle}.

	\begin{table*}
		\caption{Model Specification}
		\begin{center}
			\resizebox{\textwidth}{!}{%
					\begin{tabular}{ccccc}
						\toprule
						Model & Variant & Parameters$\downarrow$ & Input/Sensor & Output\\
						\toprule
						
						\multirow{2}{*}{TransFuser \cite{transfuser}\cite{transfuser2}} & Late Fusion & 32.64M & LiDAR, RGB, GNSS, IMU, Rotary & Waypoints, Steering, Throttle\\
						& Transformer & 66.23M & LiDAR, RGB, GNSS, IMU, Rotary & Waypoints, Steering, Throttle\\
						
						\midrule
						
						\multirow{3}{*}{DeepIPCv2} & Log. Depth & 5.91M & LiDAR, GNSS, IMU, Rotary & Waypoints, Steering, Throttle\\
						& Segmentation & 5.95M $+$14M* & LiDAR, GNSS, IMU, Rotary & Segmentation*, Waypoints, Steering, Throttle\\
						& Segmentation $+$ Log. Depth & 5.96M $+$14M* & LiDAR, GNSS, IMU, Rotary & Segmentation*, Waypoints, Steering, Throttle\\
						\bottomrule                             
					\end{tabular}
				}
			\end{center}
			\label{tab:model_compare}
			\begin{tablenotes}\small
				\item *These model variants employ PolarNet \cite{polarnet} which has total parameters of around 14 million to perform point cloud segmentation.
				\item We replicate TransFuser \cite{transfuser}\cite{transfuser2} based on the codes shared by the authors at \href{https://github.com/autonomousvision/transfuser}{https://github.com/autonomousvision/transfuser}.
				
			\end{tablenotes}
		\end{table*}

		\begin{table}[t] 
			\caption{Dataset Information}
			\begin{center}
				\resizebox{\linewidth}{!}{%
					\begin{tabular}{p{0.225\linewidth}p{0.775\linewidth}}
							\toprule
							
							Conditions & Noon, evening, night\\
							\midrule
							Total routes & 12 (train-val) and 6 (test)\\
							\midrule
							$\mathcal{N}$ Samples* & 19781 (train), 9695 (val), 29123 (test)\\
							\midrule
							Devices & WHILL model C2 (+ rotary encoder)\\
							& Velodyne LiDAR HDL-32e\\
							& Stereolabs Zed RGBD camera\\
							& U-blox Zed-F9P GNSS receiver\\
							& Witmotion HWT905 9-axis IMU sensor\\
							
							\midrule
							Object classes & None, car, bicycle, motorcycle, truck, other vehicle, person, bicyclist, motorcyclist, road, parking, sidewalk, ground, building, fence, vegetation, trunk, terrain, pole, traffic sign\\
							
							\bottomrule                             
						\end{tabular}
					}
				\end{center}
				\label{tab:data_info}
				\begin{tablenotes}\small
					\item *$\mathcal{N}$ Samples is the number of observation sets. One observation set consists of RGBD image, GNSS location, 9-axis IMU measurement, wheel's angular speed, and the level of steering and throttle.
				\end{tablenotes}
			\end{table}

		\subsection{Training}\label{subs:train}
		
		\rev{As defined in Eq. \ref{eq:mtlloss_moved}, the model is supervised using a Multi-Task Loss function.} To be more specific, we use L1 loss as in (\ref{eq:l1loss1}) in supervising waypoints prediction.
		
		\begin{equation} \label{eq:l1loss1}
			\mathcal{L}_{WP} = \frac{1}{N} \sum_{i = 1}^{N} |y_i - \hat{y}_i|,
		\end{equation}
		
		\noindent where $N$ is equal to 6 as there are three waypoints that have x,y elements in the local coordinate. Meanwhile, $y_i$ and $\hat{y}_i$ are the ground truth and the prediction of component $i$ respectively. Similarly, we also use L1 loss to supervise steering and throttle estimation formulated with (\ref{eq:l1loss2}). Keep in mind that there is no averaging process as there is only one element for each output
		
		\begin{equation} \label{eq:l1loss2}
			\mathcal{L}_{\{ST,TH\}} = |\hat{y} - y|
		\end{equation}
		
		 The model is implemented with PyTorch framework \cite{torch} and trained on NVIDIA RTX 3090 with a batch size of 10. We use Adam optimizer \cite{optim_adam} with decoupled weight decay of 0.001 \cite{adamw}. The initial learning rate is set to 0.0001 and reduced by half if the validation $\mathcal{L}_{MTL}$ is not dropping in 5 epochs in a row. The train-val process will be stopped if there is no drop on the validation $\mathcal{L}_{MTL}$ in 30 epochs in a row. 

		\subsection{Evaluation and Scoring} \label{subs:eval}
		The evaluation is conducted under three different conditions (noon, evening, night). We consider two different evaluations namely offline and online tests. In the offline test, DeepIPCv2 is deployed to predict expert driving records on the test routes. Each record has six different routes that are recorded on different days to vary the situation. Then, the performance is defined by the total metric (TM) score as in (\ref{eq:tm}).

		\begin{equation} \label{eq:tm}
			TM = MAE_{WP} + MAE_{ST} + MAE_{TH}
		\end{equation}
		
		\noindent where $MAE$ stands for mean absolute error (also known as L1 loss) which can be computed with (\ref{eq:l1loss1}) for $MAE_{WP}$ and (\ref{eq:l1loss2}) for $MAE_{ST}$ and $MAE_{TH}$. The smaller the total metric score means the better the performance. Meanwhile, in the online test, DeepIPCv2 must drive the vehicle and complete six different routes. We determine the drivability performance by counting the number of interventions and intervention time needed to prevent any collisions. The smaller the number of interventions and intervention time means the better the performance. Then, the final score for offline and online tests must be averaged as the evaluation is conducted three times.

\begin{table*}
	\caption{Multi-task Performance Score 1: Ablation and Comparative Studies}
	\begin{center}
		\resizebox{\linewidth}{!}{%
			\begin{tabular}{ccccccc}
				\toprule
				Condition&Model&Variant&Total Metric$\downarrow$&$MAE_{WP}\downarrow$&$MAE_{ST}\downarrow$&$MAE_{TH}\downarrow$\\
				\toprule 
				\multirow{5}{*}{Noon} &\multirow{2}{*}{TransFuser \cite{transfuser}\cite{transfuser2}}    &Late Fusion   &0.211 $\pm$0.007       &0.087 &0.097 &0.027 \\
				& &Transformer   &0.192 $\pm$0.006       &0.073 &0.093 &0.026 \\
				&\multirow{3}{*}{{\bf DeepIPCv2}}          &Logarithmic Depth            &0.276 $\pm$0.004 &0.116 &0.123 &0.037 \\
				& &{\bf Segmentation}   &{\bf 0.168 $\pm$0.005}       &{\bf 0.059} &{\bf 0.085} &{\bf 0.024} \\
				& &Segmentation $+$ Logarithmic Depth   &0.196 $\pm$0.007       &0.074 &0.095 &0.026 \\
				
				\midrule
				\multirow{5}{*}{Evening} &\multirow{2}{*}{TransFuser \cite{transfuser}\cite{transfuser2}}    &Late Fusion   &0.213 $\pm$0.006       &0.089 &0.097 &0.027 \\
				& &Transformer   &0.193 $\pm$0.008       &0.073 &0.094 &0.026 \\
				&\multirow{3}{*}{{\bf DeepIPCv2}}          &Logarithmic Depth            &0.281 $\pm$0.007 &0.119 &0.126 &0.036 \\
				& &{\bf Segmentation}   &{\bf 0.167 $\pm$0.006}       &{\bf 0.059} &{\bf 0.084} &{\bf 0.023} \\
				& &Segmentation $+$ Logarithmic Depth   &0.199 $\pm$0.008       &0.076 &0.097 &0.026 \\
				
				\midrule
				\multirow{5}{*}{Night} &\multirow{2}{*}{TransFuser \cite{transfuser}\cite{transfuser2}}    &Late Fusion   &0.218 $\pm$0.002       &0.090 &0.099 &0.029 \\
				& &Transformer   &0.197 $\pm$0.003       &0.075 &0.094 &0.028 \\
				&\multirow{3}{*}{{\bf DeepIPCv2}}          &Logarithmic Depth            &0.278 $\pm$0.005 &0.115 &0.125 &0.038 \\
				& &{\bf Segmentation}   &{\bf 0.170 $\pm$0.002}       &{\bf 0.059} &{\bf 0.086} &{\bf 0.026} \\
				& &Segmentation $+$ Logarithmic Depth   &0.198 $\pm$0.004       &0.072 &0.097 &0.028 \\
				\bottomrule
			\end{tabular}
		}
	\end{center}
	\label{tab:mtl_result}
\end{table*}

		\subsubsection{Ablation and Comparative Studies}
		As mentioned in Subsection \ref{subs:model}, we create two model variants for the ablation study. The first variant only takes the logarithmic depth point clouds while the second variant only takes the segmented point clouds. Then, after obtaining the best variant, we conduct another ablation study by comparing it with other variants that perceive the surroundings using only one perspective, front or BEV. Furthermore, we also conduct a comparative study by replicating TransFuser \cite{transfuser}\cite{transfuser2} to compare with. Briefly, TransFuser is a camera-LiDAR fusion model that takes an RGB image and a set of point clouds. It perceives the environment from two different perspectives where the front view information is given by the RGB camera and the BEV information is given by the LiDAR. TransFuser fuses RGB images and LiDAR point clouds features using several transformer modules. We also replicate its variant called late fusion, where the features are fused with a simple element-wise summation. The specification of DeepIPCv2 and TransFuser variants can be seen in Table \ref{tab:model_compare}. \rev{Instead of fluctuating runtime measurements due to the variability of runtime on GPU resources, we utilize Total Parameters as the deterministic metric for computational efficiency. DeepIPCv2 is approximately $3.3\times$ smaller than the TransFuser baseline (5.95M+14M vs 66.23M). This significant reduction in model size inherently lowers memory bandwidth requirements and computational cost, ensuring the system easily meets the real-time requirements for the vehicle's operating speed of 1.25 m/s.}

		\section{Result and Discussion}

		\subsection{Offline Test}
		An offline test is used to measure how good the model is in mimicking an expert by predicting several driving records made for testing purposes. We measure the model performance by calculating the MAE on waypoints prediction and navigational control estimation together with the total metric (TM) score as explained in Subsection \ref{subs:eval}. Since there are three driving records for each condition, the final score for each condition is averaged from all inference results. %
		
		\subsubsection{Comparative Analysis of Sensor Modalities}
		Based on Table \ref{tab:mtl_result}, the DeepIPCv2 variant that only takes segmented point clouds achieves the best performance by having the lowest TM score in all conditions. The other two DeepIPCv2 variants that take logarithmic depth point clouds fall behind and the depth-only variant performs the worst. This pattern shows that processing logarithmic depth reduces overall model performance due to conflicting features between the segmentation map and the depth map extracted by the encoders. When combined with the clean class-based features of the segmentation map, this may confuse the model rather than help it. However, this hypothesis needs to be justified further by applying different encoder architectures to process the projected point clouds. Meanwhile, amongst TransFuser variants, the variant that employs transformer modules to fuse image and point cloud features achieves a better performance than the variant that only uses a simple element-wise summation. The transformer modules can improve the model's reasoning as it understands the relationship between the front view and BEV perspectives.

		\begin{table*}[t]
			\caption{Multi-task Performance Score 2: The Importance of Multiple Perspectives}
			\begin{center}
				\resizebox{0.65\linewidth}{!}{%
					\begin{tabular}{cccccc}
						\toprule
						Condition&Perspective&Total Metric$\downarrow$&$MAE_{WP}\downarrow$&$MAE_{ST}\downarrow$&$MAE_{TH}\downarrow$\\
						\toprule 
						\multirow{3}{*}{Noon} &Front   &0.258 $\pm$0.006       &0.062 &0.173 &{\bf 0.023} \\
						&BEV   &0.171 $\pm$0.006       &0.063 &{\bf 0.084} &0.024 \\
						&{\bf Front $+$ BEV}   &{\bf 0.168 $\pm$0.005}       &{\bf 0.059} &0.085 &0.024 \\
						
						\midrule
						\multirow{3}{*}{Evening} &Front   &0.258 $\pm$0.014       &0.063 &0.173 &{\bf 0.022} \\
						&BEV   &0.171 $\pm$0.005       &0.062 &0.085 &0.023 \\
						&{\bf Front $+$ BEV}   &{\bf 0.167 $\pm$0.006}       &{\bf 0.059} &{\bf 0.084} &0.023 \\
						
						\midrule
						\multirow{3}{*}{Night} &Front   &0.263 $\pm$0.005       &0.061 &0.177 &{\bf 0.025} \\
						&BEV   &0.174 $\pm$0.004       &0.062 &{\bf 0.086} &0.026 \\
						&{\bf Front $+$ BEV}   &{\bf 0.170 $\pm$0.002}       &{\bf 0.059} &{\bf 0.086} &0.026 \\
						\bottomrule
					\end{tabular}
				}
			\end{center}
			\label{tab:mtl_result2}
		\end{table*}

		In the comparison across different conditions, the total metric scores for both TransFuser variants consistently become higher from predicting noon records to night records. Although the gap is not too far from one another, this pattern shows that TransFuser which relies on RGB images gets a performance drop when the illumination condition is poor. This result is as expected since the RGB camera is sensitive to illumination changes. Unlike TransFuser, DeepIPCv2 is more robust against poor illumination conditions as it only relies on LiDAR to perceive the environment. However, there is no clear pattern amongst DeepIPCv2 variants as their performance differs on every condition. DeepIPCv2 performance is more affected by road situations rather than illumination. This is supported by the fact that DeepIPCv2 variants have the best performance at night when there is less traffic.

		\subsubsection{More Perspectives, Better Reasoning}
		
		To understand the importance of perceiving from multiple perspectives of view, we also conduct an extensive ablation study by creating two more DeepIPCv2 variants that take one perspective of view. Hence, the model can only perceive the environment based on the front-view perspective or the top-view/bird's eye view (BEV) perspective. We develop these variants based on DeepIPCv2 variant that takes the point cloud segmentation map. Table \ref{tab:mtl_result2} shows that the model variant that perceives the environment from both front and BEV perspectives has the lowest total metric score meaning that it achieves the best performance compared to the variants which only use one perspective. This result is in line with the findings in our previous work \cite{deepipc}\cite{oskar_tiv} and strengthens the importance of perceiving from multiple perspectives of view.
		
		To be more detailed, DeepIPCv2 variant that perceives from the front-view perspective has the worst performance as its steering estimation is heavily affected by the absence of BEV perception features. Without these features, the controller module faces difficulties in estimating the steering angle which lies in the BEV coordinate. Meanwhile, DeepIPCv2 variant that perceives from the BEV perspective is slightly behind the best variant. Although it has a comparable performance in steering angle and throttle level estimation, this variant fails to estimate waypoints properly as predicting future vehicle position needs the combination of both front and BEV perception features that consider more aspects of driving. Therefore, judging from this result and analysis, we pick the DeepIPCv2 variant that only takes segmented point clouds and perceives the environment from multiple perspectives of view for further comparison in the online test. Meanwhile, we pick TransFuser variant that employs transformer modules as the comparator to study the importance of data modality and representation in performing real-world autonomous driving.

		\begin{table}
			\caption{Driving Performance: Drivability Score}
			\begin{center}
				\resizebox{\linewidth}{!}{%
					\begin{tabular}{cccc}
						\toprule
						\multirow{2}{*}{Condition} & \multirow{2}{*}{Model} & \multicolumn{2}{c}{Intervention$\downarrow$}\\
						& & Count & Time (secs)\\
						\toprule 
						\multirow{2}{*}{Noon} &TransFuser \cite{transfuser}\cite{transfuser2}             &1.389 $\pm$0.208       & 3.537 $\pm$0.648\\
						&{\bf DeepIPCv2}                            &{\bf 1.000 $\pm$0.236} &{\bf 2.389 $\pm$0.831}\\
						\midrule
						\multirow{2}{*}{Evening} &TransFuser \cite{transfuser}\cite{transfuser2}          &1.222 $\pm$0.079       & 3.093 $\pm$0.457\\
						&{\bf DeepIPCv2}  				           &{\bf 0.944 $\pm$0.157} 	     &{\bf 2.407 $\pm$0.466}\\
						\midrule
						\multirow{2}{*}{Night} &TransFuser \cite{transfuser}\cite{transfuser2}          &1.889 $\pm$0.283       & 4.556 $\pm$0.181\\
						&{\bf DeepIPCv2}  				           &{\bf 0.667 $\pm$0.136} 	     &{\bf 1.870 $\pm$0.340}\\
						\bottomrule
					\end{tabular}
				}
			\end{center}
			\label{tab:drive_result} 
		\end{table}

		\subsection{Online Test}

		An online test is made for evaluating the drivability performance of the model after imitating expert behavior in driving a vehicle during the training process. The best variant of DeepIPCv2 and TransFuser are evaluated by being deployed for automated driving in real environments. Each model must be able to handle various situations and conditions when driving a vehicle from the starting point to the finish point by following a set of route points. Similar to the offline test, we conduct the test three times on six different routes for each condition. However, the drivability performance is justified by the number of interventions and how long the interventions are. The best performance is determined by the lowest number of interventions and the shortest time of intervention. Keep in mind that the result of the online test may not be in line with the result of the offline test. This is because any decisions made on every observation state in the online test will affect the next observation state. Meanwhile, in the offline test, although the model makes a wrong prediction, it will not affect anything as the observation is already fixed in the driving records. Table \ref{tab:drive_result} shows a clear pattern for each model when driving under different conditions.

		\begin{figure*}
			\begin{center}
				\includegraphics[width=\linewidth]{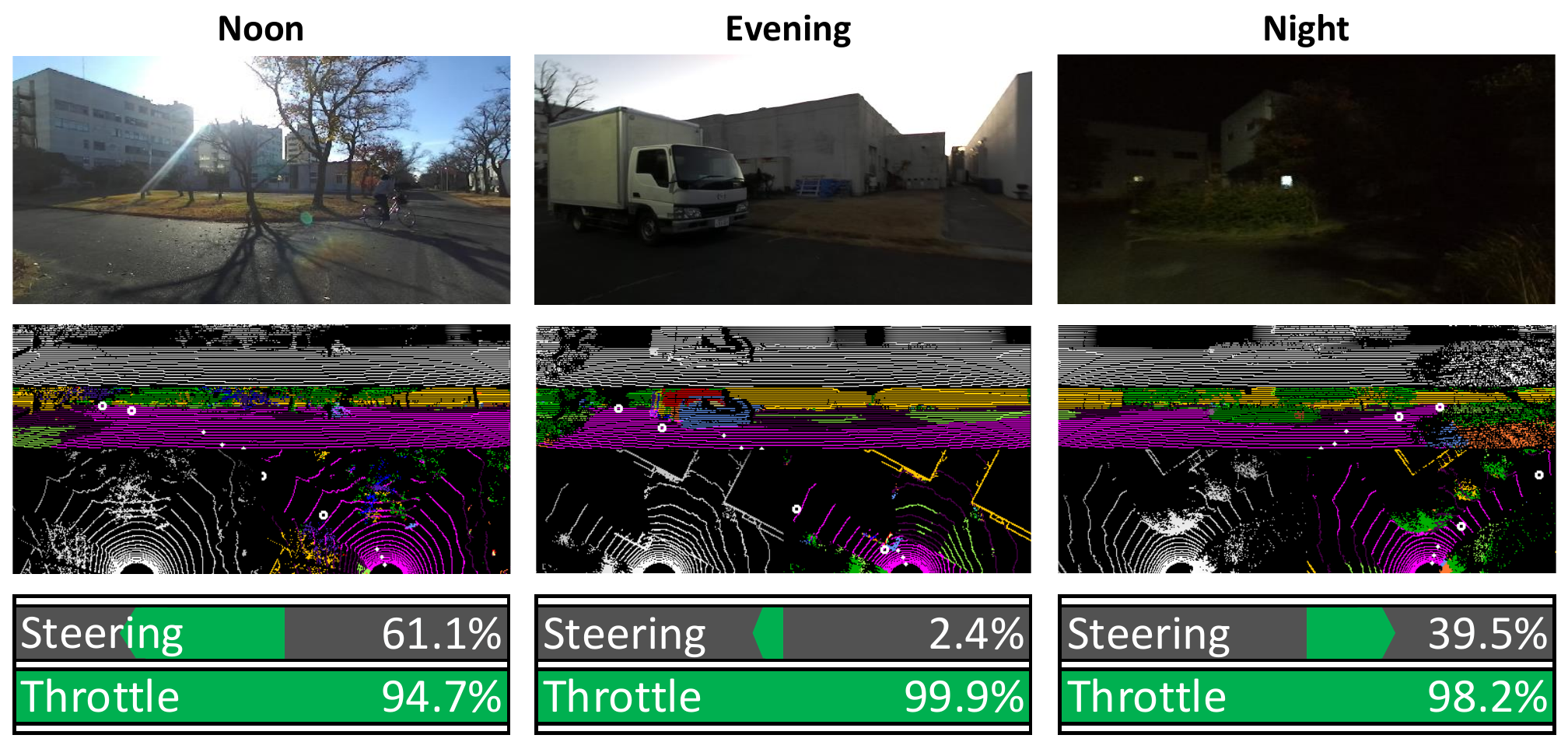}
				
			\end{center}
			\caption{Driving footage. Noon: DeepIPCv2 makes a left turn following two route points located on the left side of the vehicle. Evening: A moment when we are going to intervene in DeepIPCv2 as it fails to make a hard left turn to compromise a narrowing path caused by a stopping truck. Night: DeepIPCv2 makes a right turn following two route points while maintaining the distance from the road boundaries. To be noted, the RGB images are only for record purposes. We share some of the driving records with a playback speed of 5$\times$, which includes the comparison with TransFuser \cite{transfuser}\cite{transfuser2} at \href{https://youtu.be/IsZ1HP5QjWc}{https://youtu.be/IsZ1HP5QjWc}.}
			\label{fig:driving_rec}
		\end{figure*}

		\subsubsection{Robustness to Illumination}
		As shown in Fig. \ref{fig:driving_rec}, DeepIPCv2 that is not affected by illumination conditions achieves the best drivability at night when there is not so much traffic on the road. Then, it has lower performance at noon and in the evening as it is affected by denser traffic on the road. Meanwhile, as for TransFuser which relies on the RGB camera, better performance is achieved when the model drives at noon and in the evening. Thanks to enough light illumination, the RGB camera can capture an image clearly so that TransFuser can maintain its drivability. However, TransFuser performance is degraded when driving at night as it fails to capture the information due to poor illumination conditions. Therefore, as the perception module cannot extract useful features, the controller module fails to estimate navigational control properly. 
		
		\subsubsection{Impact of Semantic Representation}
		In all conditions, DeepIPCv2 has the best performance based on the lowest intervention count and intervention time compared to TransFuser. This means that a set of segmented point clouds projected into two different perspectives of view contains more valuable information than a combination of a raw RGB image and a 2-bin point cloud histogram. By projecting the segmented point clouds to form image-like arrays that contain a unique class on each layer, the model has a better scene understanding capability as it can distinguish traversable and non-traversable areas clearly and lead to better driving performance. This also shows that data representation (e.g. segmented and projected point clouds) is matter and more meaningful than a combination of some data modalities (e.g. RGB images and LiDAR point clouds) but still in their raw form or improperly pre-processed. 
		
		\subsubsection{Limitations and Failure Analysis}
		\rev{We observed failure cases in scenarios with dynamic occlusions, such as pedestrians suddenly stepping out from behind parked trucks. In these instances, although the LiDAR perception module correctly segments the obstacle in the BEV map, the control policy occasionally exhibits a response lag. This latency is primarily attributed to the damping characteristics of the PID controller, which is tuned to prioritize smooth trajectory tracking and passenger comfort over the high-jerk maneuvers required for emergency collision avoidance. Additionally, the heuristic command generation, which relies on fixed lateral thresholds ($\pm$4m), can be rigid on non-standard road curvatures. On unstructured paths where the road geometry does not align perfectly with these hard thresholds, the system may occasionally misclassify the high-level navigational command (e.g., confusing a sharp curve with a navigational turn), leading to suboptimal steering inferences.}

		\section{Conclusion}
		We propose DeepIPCv2 which perceives the environment using LiDAR for more robust drivability, either in good and poor illumintation conditions. DeepIPCv2 is evaluated by predicting a set of driving records and performing real automated driving on an outdoor vehicle. To better justify its performance, we conduct both ablation and comparative studies with other models and variants under different conditions to vary the situations.

		Based on the experimental results, we disclose that using LiDAR to perceive the environment increases the model's robustness. Unlike an RGB camera, LiDAR is not affected by poor illumination conditions. Thus, the perception module can provide stable features to the controller module in estimating navigational control properly. Therefore, the model can maintain its drivability even when driving at night (poor illumination conditions). Meanwhile, the performance of a camera-powered model drops as it fails to perceive the surrounding area. Then, we also disclose that perceiving the environment with segmented point clouds that are projected into multi-view perspectives is better than a combination of raw RGB images and 2-bin histogram point clouds. As the model has better reasoning, the overall driving performance is increased, especially when driving at night with severe illumination conditions. \rev{However, the selection of the LiDAR segmentation model must be carried out carefully considering the trade-off between latency and performance for real-time inference.}

		\rev{We acknowledge that the current evaluation is limited to a single campus environment, which may constrain the model's generalization to unseen geographic locations.} In the future, more crowded environments and more adversarial scenarios can be used to test the model further. As the driving challenges increased, the model also needed to be enhanced with a sensor that can detect event changes such as a DVS camera to improve the perception. Then, better encoders and fusion techniques might be needed to handle different data modalities and representations. \rev{Furthermore, to optimize the model for embedded automotive hardware, we plan to explore asymmetric kernel decomposition (e.g., replacing $5\times5$ convolutions with $1\times5$ and $5\times1$ layers). Finally, we intend to replace the static PID gains with an adaptive tuning mechanism based on evolutionary algorithms to dynamically adjust to varying vehicle dynamics.}

\section*{Acknowledgments}
This publication is funded by the Indonesian Endowment Fund for Education (LPDP) on behalf of the Indonesian Ministry of Higher Education, Science and Technology and managed under the EQUITY Program (Contract Number: 4301/B3/DT.03.08/2025 and 10107/UN1.P/Dit-Keu/HK.08.00/2025).


\bibliographystyle{IEEEtran}
\bibliography{references}

\begin{thebibliography}{10}
\providecommand{\url}[1]{#1}
\csname url@samestyle\endcsname
\providecommand{\newblock}{\relax}
\providecommand{\bibinfo}[2]{#2}
\providecommand{\BIBentrySTDinterwordspacing}{\spaceskip=0pt\relax}
\providecommand{\BIBentryALTinterwordstretchfactor}{4}
\providecommand{\BIBentryALTinterwordspacing}{\spaceskip=\fontdimen2\font plus
\BIBentryALTinterwordstretchfactor\fontdimen3\font minus
  \fontdimen4\font\relax}
\providecommand{\BIBforeignlanguage}[2]{{%
\expandafter\ifx\csname l@#1\endcsname\relax
\typeout{** WARNING: IEEEtran.bst: No hyphenation pattern has been}%
\typeout{** loaded for the language `#1'. Using the pattern for}%
\typeout{** the default language instead.}%
\else
\language=\csname l@#1\endcsname
\fi
#2}}
\providecommand{\BIBdecl}{\relax}
\BIBdecl

\bibitem{av_vision0}
J.~Horgan, C.~Hughes, J.~McDonald, and S.~Yogamani, ``Vision-based driver
  assistance systems: Survey, taxonomy and advances,'' in \emph{Proc. IEEE
  Intell. Transp. Syst. Conf. (ITSC)}, Gran Canaria, Spain, Sep. 2015, pp.
  2032--2039.

\bibitem{avisionx}
Y.~Cui, R.~Chen, W.~Chu, L.~Chen, D.~Tian, Y.~Li, and D.~Cao, ``Deep learning
  for image and point cloud fusion in autonomous driving: A review,''
  \emph{IEEE Trans. Intell. Transp. Syst.}, vol.~23, no.~2, pp. 722--739, Feb.
  2022.

\bibitem{avisiony}
K.~Muhammad, T.~Hussain, H.~Ullah, J.~D. Ser, M.~Rezaei, N.~Kumar, M.~Hijji,
  P.~Bellavista, and V.~H.~C. de~Albuquerque, ``Vision-based semantic
  segmentation in scene understanding for autonomous driving: Recent
  achievements, challenges, and outlooks,'' \emph{IEEE Trans. Intell. Transp.
  Syst.}, vol.~23, no.~12, pp. 22\,694--22\,715, Dec. 2022.

\bibitem{avisionz}
D.~Omeiza, H.~Webb, M.~Jirotka, and L.~Kunze, ``Explanations in autonomous
  driving: A survey,'' \emph{IEEE Trans. Intell. Transp. Syst.}, vol.~23,
  no.~8, pp. 10\,142--10\,162, Aug. 2022.

\bibitem{obj_det}
G.~Adam, V.~Chitalia, N.~Simha, A.~Ismail, S.~Kulkarni, V.~Narayan, and
  M.~Schulze, ``Robustness and deployability of deep object detectors in
  autonomous driving,'' in \emph{Proc. IEEE Intell. Transp. Syst. Conf.
  (ITSC)}, Auckland, New Zealand, Oct. 2019, pp. 4128--4133.

\bibitem{sem_seg}
C.~Wang and N.~Aouf, ``Fusion attention network for autonomous cars semantic
  segmentation,'' in \emph{Proc. IEEE Intell. Veh. Symp. (IV)}, Aachen,
  Germany, Jul. 2022, pp. 1525--1530.

\bibitem{matsuzaki}
S.~Matsuzaki, H.~Masuzawa, and J.~Miura, ``Multi-source soft pseudo-label
  learning with domain similarity-based weighting for semantic segmentation,''
  in \emph{Proc. IEEE/RSJ Inter. Conf. Intell. Robots and Syst. (IROS)},
  Detroit, USA, Oct. 2023, pp. 5852--5857.

\bibitem{segx}
O.~Natan, D.~U.~K. Putri, and A.~Dharmawan, ``Deep learning-based weld spot
  segmentation using modified {UNet} with various convolutional blocks,''
  \emph{ICIC Express Letters Part B: Applications}, vol.~12, no.~12, pp.
  1169--1176, Dec. 2021.

\bibitem{dep_est}
A.~Gurram, A.~F. Tuna, F.~Shen, O.~Urfalioglu, and A.~M. López, ``Monocular
  depth estimation through virtual-world supervision and real-world {SfM}
  self-supervision,'' \emph{IEEE Trans. Intell. Transp. Syst.}, vol.~23, no.~8,
  pp. 12\,738--12\,751, Aug. 2022.

\bibitem{seg_seq}
H.-k. Chiu, E.~Adeli, and J.~C. Niebles, ``Segmenting the future,'' \emph{IEEE
  Robot. and Autom. Lett.}, vol.~5, no.~3, pp. 4202--4209, Jul. 2020.

\bibitem{mtl}
T.-J. Song, J.~Jeong, and J.-H. Kim, ``End-to-end real-time obstacle detection
  network for safe self-driving via multi-task learning,'' \emph{IEEE Trans.
  Intell. Transp. Syst.}, vol.~23, no.~9, pp. 16\,318--16\,329, Sep. 2022.

\bibitem{mtl_tii}
R.~Xie, C.~Li, X.~Zhou, H.~Chen, and Z.~Dong, ``Differentially private
  federated learning for multitask objective recognition,'' \emph{IEEE Trans.
  Industr. Inform.}, vol.~20, no.~5, pp. 7269--7281, Feb. 2024.

\bibitem{sensor_fuse0}
S.~Xu, D.~Zhou, J.~Fang, J.~Yin, Z.~Bin, and L.~Zhang, ``{FusionPainting}:
  Multimodal fusion with adaptive attention for {3D} object detection,'' in
  \emph{Proc. IEEE Intell. Transp. Syst. Conf. (ITSC)}, Indianapolis, USA, Oct.
  2021, pp. 3047--3054.

\bibitem{rgbdvs}
O.~Natan and J.~Miura, ``Semantic segmentation and depth estimation with {RGB}
  and {DVS} sensor fusion for multi-view driving perception,'' in \emph{Proc.
  Asian Conf. Pattern Recog. (ACPR)}, Jeju Island, South Korea, Nov. 2021, pp.
  352--365.

\bibitem{rgb_d}
L.~Sun, K.~Yang, X.~Hu, W.~Hu, and K.~Wang, ``Real-time fusion network for
  {RGB-D} semantic segmentation incorporating unexpected obstacle detection for
  road-driving images,'' \emph{IEEE Robot. and Autom. Lett.}, vol.~5, no.~4,
  pp. 5558--5565, Oct. 2020.

\bibitem{senfus_tie}
X.~Hao, Y.~Xia, H.~Yang, and Z.~Zuo, ``Asynchronous information fusion in
  intelligent driving systems for target tracking using cameras and radars,''
  \emph{IEEE Trans. Industr. Electron.}, vol.~70, no.~3, pp. 2708--2717, Apr.
  2023.

\bibitem{deepipc}
O.~Natan and J.~Miura, ``{DeepIPC}: Deeply integrated perception and control
  for an autonomous vehicle in real environments,'' \emph{IEEE Access},
  vol.~12, pp. 49\,590--49\,601, Apr. 2024.

\bibitem{lidar1}
Z.~He, X.~Fan, Y.~Peng, Z.~Shen, J.~Jiao, and M.~Liu, ``{EmPointMovSeg}: Sparse
  tensor-based moving-object segmentation in {3-D LiDAR} point clouds for
  autonomous driving-embedded system,'' \emph{IEEE Trans. Comput.-Aided Des.
  Integr. Circuits Syst.}, vol.~42, no.~1, pp. 41--53, Jan. 2023.

\bibitem{lidar3}
S.~Zhou, H.~Xu, G.~Zhang, T.~Ma, and Y.~Yang, ``Leveraging deep convolutional
  neural networks pre-trained on autonomous driving data for vehicle detection
  from roadside {LiDAR} data,'' \emph{IEEE Trans. Intell. Transp. Syst.},
  vol.~23, no.~11, pp. 22\,367--22\,377, Nov. 2022.

\bibitem{lidar4}
G.~Xian, C.~Ji, L.~Zhou, G.~Chen, J.~Zhang, B.~Li, X.~Xue, and J.~Pu,
  ``Location-guided {LiDAR}-based panoptic segmentation for autonomous
  driving,'' \emph{IEEE Trans. Intell. Veh.}, vol.~8, no.~2, pp. 1473--1483,
  Feb. 2023.

\bibitem{oskar_tiv}
O.~Natan and J.~Miura, ``End-to-end autonomous driving with semantic depth
  cloud mapping and multi-agent,'' \emph{IEEE Trans. Intell. Veh.}, vol.~8,
  no.~1, pp. 557--571, Jan. 2022.

\bibitem{transfuser2}
K.~Chitta, A.~Prakash, B.~Jaeger, Z.~Yu, K.~Renz, and A.~Geiger,
  ``{TransFuser}: Imitation with transformer-based sensor fusion for autonomous
  driving,'' \emph{IEEE Trans. Pattern Anal. Mach. Intell.}, vol.~45, no.~11,
  pp. 12\,878 -- 12\,895, Nov. 2023.

\bibitem{polarnet}
Y.~Zhang, Z.~Zhou, P.~David, X.~Yue, Z.~Xi, B.~Gong, and H.~Foroosh,
  ``{PolarNet}: An improved grid representation for online {LiDAR} point clouds
  semantic segmentation,'' in \emph{Proc. IEEE/CVF Conf. Comput. Vision and
  Pattern Recog. (CVPR)}, Seattle, USA, Jun. 2020, pp. 9598--9607.

\bibitem{robust_lidar}
R.~W. Wolcott and R.~M. Eustice, ``Robust {LiDAR} localization using
  multiresolution {Gaussian} mixture maps for autonomous driving,'' \emph{Int.
  J. Rob. Res.}, vol.~36, no.~3, pp. 292--319, Apr. 2017.

\bibitem{lidar2}
S.~McCrae and A.~Zakhor, ``{3D} object detection for autonomous driving using
  temporal {LiDAR} data,'' in \emph{Proc. Inter. Conf. Image Processing
  (ICIP)}, Abu Dhabi, UAE, Oct. 2020, pp. 2661--2665.

\bibitem{lidar_tie}
H.~Shen, Q.~Zong, B.~Tian, X.~Zhang, and H.~Lu, ``{PGO-LIOM}: Tightly coupled
  {LiDAR}-inertial odometry and mapping via parallel and gradient-free
  optimization,'' \emph{IEEE Trans. Industr. Electron.}, vol.~70, no.~11, pp.
  11\,453--11\,463, Dec. 2023.

\bibitem{pt_cloud_av0}
Y.~Li, L.~Ma, Z.~Zhong, F.~Liu, M.~A. Chapman, D.~Cao, and J.~Li, ``Deep
  learning for {LiDAR} point clouds in autonomous driving: A review,''
  \emph{IEEE Trans. Neural Networks and Learning Syst.}, vol.~32, no.~8, pp.
  3412--3432, Aug. 2021.

\bibitem{pt_cloud_av1}
Y.~Li and J.~Ibanez-Guzman, ``{LiDAR} for autonomous driving: The principles,
  challenges, and trends for automotive {LiDAR} and perception systems,''
  \emph{IEEE Signal Process. Mag.}, vol.~37, no.~4, pp. 50--61, Jul. 2020.

\bibitem{lidar_tii}
X.~Sun, M.~Wang, J.~Du, Y.~Sun, S.~S. Cheng, and W.~Xie, ``A task-driven
  scene-aware {LiDAR} point cloud coding framework for autonomous vehicles,''
  \emph{IEEE Trans. Industr. Inform.}, vol.~19, no.~8, pp. 8731--8742, Nov.
  2023.

\bibitem{kitti}
J.~Behley, M.~Garbade, A.~Milioto, J.~Quenzel, S.~Behnke, J.~Gall, and
  C.~Stachniss, ``{Towards 3D LiDAR-based semantic scene understanding of 3D
  point cloud sequences: The SemanticKITTI Dataset},'' \emph{The International
  Journal on Robotics Research}, vol.~40, no. 8-9, pp. 959--967, Apr. 2021.

\bibitem{2dpass}
X.~Yan, J.~Gao, C.~Zheng, C.~Zheng, R.~Zhang, S.~Cui, and Z.~Li, ``{2DPASS: 2D}
  priors assisted semantic segmentation on {LiDAR} point clouds,'' in
  \emph{Proc. European Conf. Comput. Vision (ECCV)}, Tel Aviv, Israel, Oct.
  2022, pp. 677--695.

\bibitem{pvkd}
Y.~Hou, X.~Zhu, Y.~Ma, C.~C. Loy, and Y.~Li, ``Point-to-voxel knowledge
  distillation for {LiDAR} semantic segmentation,'' in \emph{Proc. IEEE/CVF
  Conf. Comput. Vision and Pattern Recog. (CVPR)}, New Orleans, USA, Jun. 2022,
  pp. 8469--8478.

\bibitem{pcbev}
S.~Qiu, X.~Li, X.~Xue, and J.~Pu, ``{PC-BEV}: An efficient polar-cartesian
  {BEV} fusion framework for {LiDAR} semantic segmentation,'' \emph{Proceedings
  of the AAAI Conference on Artificial Intelligence}, vol.~39, no.~6, pp.
  6612--6620, Apr. 2025.

\bibitem{e2e_av1}
A.~Tampuu, T.~Matiisen, M.~Semikin, D.~Fishman, and N.~Muhammad, ``A survey of
  end-to-end driving: Architectures and training methods,'' \emph{IEEE Trans.
  Neural Networks and Learning Syst.}, vol.~33, no.~4, pp. 1364--1384, Apr.
  2022.

\bibitem{e2e_ava}
S.~Teng, L.~Chen, Y.~Ai, Y.~Zhou, Z.~Xuanyuan, and X.~Hu, ``Hierarchical
  interpretable imitation learning for end-to-end autonomous driving,''
  \emph{IEEE Trans. Intell. Veh.}, vol.~8, no.~1, pp. 673--683, Jan. 2023.

\bibitem{ishihara}
K.~Ishihara, A.~Kanervisto, J.~Miura, and V.~Hautamaki, ``Multi-task learning
  with attention for end-to-end autonomous driving,'' in \emph{Proc. IEEE/CVF
  Conf. Comput. Vision and Pattern Recog. Workshops (CVPRW)}, Nashville, USA,
  Jun. 2021, pp. 2896--2905.

\bibitem{mtl_a}
E.~Kargar and V.~Kyrki, ``Increasing the efficiency of policy learning for
  autonomous vehicles by multi-task representation learning,'' \emph{IEEE
  Trans. Intell. Veh.}, vol.~7, no.~3, pp. 701--710, Sep. 2022.

\bibitem{neat}
K.~Chitta, A.~Prakash, and A.~Geiger, ``{NEAT}: Neural attention fields for
  end-to-end autonomous driving,'' in \emph{Proc. IEEE/CVF Inter. Conf. Comput.
  Vision (ICCV)}, Montreal, Canada, Oct. 2021, pp. 15\,773--15\,783.

\bibitem{transfuser}
A.~Prakash, K.~Chitta, and A.~Geiger, ``Multi-modal fusion transformer for
  end-to-end autonomous driving,'' in \emph{Proc. IEEE/CVF Conf. Comput. Vision
  and Pattern Recog. (CVPR)}, Nashville, USA, Jun. 2021, pp. 7073--7083.

\bibitem{2bin_transfuser}
N.~Rhinehart, R.~Mcallister, K.~Kitani, and S.~Levine, ``{PRECOG}: Prediction
  conditioned on goals in visual multi-agent settings,'' in \emph{Proc.
  IEEE/CVF Inter. Conf. Comput. Vision (ICCV)}, Seoul, South Korea, Nov. 2019,
  pp. 2821--2830.

\bibitem{atrous_conv}
L.-C. Chen, G.~Papandreou, I.~Kokkinos, K.~Murphy, and A.~L. Yuille,
  ``{DeepLab}: Semantic image segmentation with deep convolutional nets, atrous
  convolution, and fully connected {CRFs},'' \emph{IEEE Trans. Pattern Anal.
  Mach. Intell.}, vol.~40, no.~4, pp. 834--848, Apr. 2018.

\bibitem{gru}
K.~Cho, B.~van Merrienboer, D.~Bahdanau, and Y.~Bengio, ``On the properties of
  neural machine translation: Encoder-decoder approaches,'' in \emph{Proc.
  Workshop Syntax, Semantics and Structure in Statistical Translation (SSST)},
  Doha, Qatar, Oct. 2014, pp. 103--111.

\bibitem{mgn}
O.~Natan and J.~Miura, ``Towards compact autonomous driving perception with
  balanced learning and multi-sensor fusion,'' \emph{IEEE Trans. Intell.
  Transp. Syst.}, vol.~23, no.~9, pp. 16\,249--16\,266, Sep. 2022.

\bibitem{huang_model}
Z.~Huang, C.~Lv, Y.~Xing, and J.~Wu, ``Multi-modal sensor fusion-based deep
  neural network for end-to-end autonomous driving with scene understanding,''
  \emph{IEEE Sensors J.}, vol.~21, no.~10, pp. 11\,781--11\,790, May 2021.

\bibitem{imit4}
F.~S. Acerbo, M.~Alirczaei, H.~Van Der~Auweraer, and T.~D. Son, ``Safe
  imitation learning on real-life highway data for human-like autonomous
  driving,'' in \emph{Proc. IEEE Intell. Transp. Syst. Conf. (ITSC)},
  Indianapolis, USA, Sep. 2021, pp. 3903--3908.

\bibitem{imita}
D.~Sun, Q.~Liao, and A.~Loutfi, ``Type-2 fuzzy model-based movement primitives
  for imitation learning,'' \emph{IEEE Trans. Robot.}, vol.~38, no.~4, pp.
  2462--2480, Aug. 2022.

\bibitem{hidehito}
H.~Fujiishi, T.~Kobayashi, and K.~Sugimoto, ``Safe and efficient imitation
  learning by clarification of experienced latent space,'' \emph{Adv. Robot.},
  vol.~35, no.~16, pp. 1012--1027, Jul. 2021.

\bibitem{imitb}
M.~Alibeigi, M.~N. Ahmadabadi, and B.~N. Araabi, ``A fast, robust, and
  incremental model for learning high-level concepts from human motions by
  imitation,'' \emph{IEEE Trans. Robot.}, vol.~33, no.~1, pp. 153--168, Feb.
  2017.

\bibitem{torch}
A.~Paszke, S.~Gross, F.~Massa, A.~Lerer, J.~Bradbury, G.~Chanan, T.~Killeen,
  Z.~Lin, N.~Gimelshein, L.~Antiga, A.~Desmaison, A.~Kopf, E.~Yang, Z.~DeVito,
  M.~Raison, A.~Tejani, S.~Chilamkurthy, B.~Steiner, L.~Fang, J.~Bai, and
  S.~Chintala, ``{PyTorch}: An imperative style, high performance deep learning
  library,'' in \emph{Proc. Inter. Conf. Neural Information Processing Syst.
  (NIPS)}, Vancouver, Canada, Dec. 2019, pp. 8024--8035.

\bibitem{optim_adam}
D.~P. Kingma and J.~Ba, ``Adam: A method for stochastic optimization,'' in
  \emph{Proc. Inter. Conf. Learning Representations (ICLR)}, San Diego, USA,
  May 2015.

\bibitem{adamw}
I.~Loshchilov and F.~Hutter, ``Decoupled weight decay regularization,'' in
  \emph{Proc. Inter. Conf. Learning Representations (ICLR)}, New Orleans, USA,
  May 2019, pp. 1--10.

\end{thebibliography}
%


\newpage
\begin{IEEEbiography}[{\includegraphics[width=1in,height=1.25in,clip,keepaspectratio]{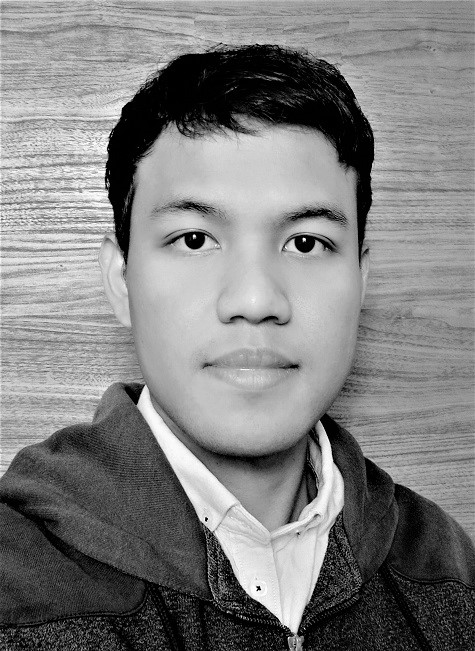}}]{Oskar Natan}
	(Member, IEEE) received his B.A.Sc. degree in Electronics Engineering and M.Eng. degree in Electrical Engineering from Politeknik Elektronika Negeri Surabaya, Indonesia, in 2017 and 2019, respectively. In 2023, he received his Ph.D.(Eng.) degree in Computer Science and Engineering from Toyohashi University of Technology, Japan. Since January 2020, he has been affiliated with the Department of Computer Science and Electronics, Universitas Gadjah Mada, Indonesia, first as a Lecturer and currently serves as an Assistant Professor. He has been serving as a reviewer/TPC member for some reputable journals and conferences. His research interests lie in the fields of deep learning, sensor fusion, hardware acceleration, and end-to-end systems.  
\end{IEEEbiography}




\begin{IEEEbiography}[{\includegraphics[width=1in,height=1.25in,clip,keepaspectratio]{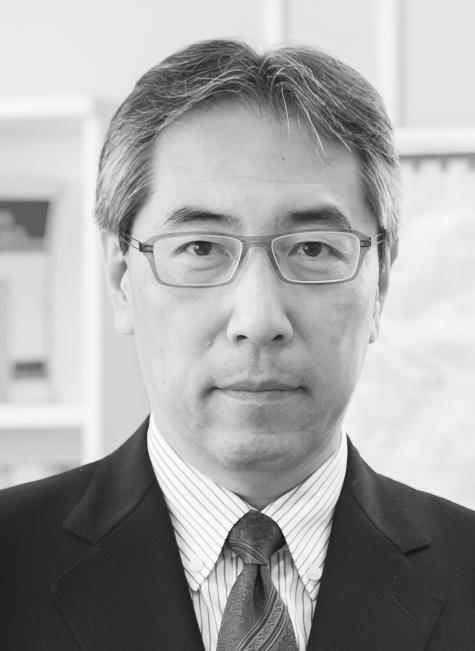}}]{Jun Miura}
	(Member, IEEE) received his B.Eng. degree in Mechanical Engineering and his M.Eng. and Dr.Eng. degrees in Information Engineering from the University of Tokyo, Japan, in 1984, 1986, and 1989, respectively. From 1989 to 2007, he was with the Department of Computer-controlled Mechanical Systems, Osaka University, Japan, first as a Research Associate and later as an Associate Professor. From March 1994 to February 1995, he served as a Visiting Scientist at the Department of Computer Science, Carnegie Mellon University, USA. In 2007, he became a Professor at the Department of Computer Science and Engineering, Toyohashi University of Technology, Japan, where he remains to the present. To date, he has received plenty of awards and authored or co-authored more than 265 peer-reviewed scientific articles in the field of robotics and autonomous systems in internationally reputable journals and conferences.
\end{IEEEbiography}

\vfill
\EOD

\end{document}